\documentclass{article}

\usepackage[final]{neurips_2025}

\usepackage[utf8]{inputenc} 
\usepackage[T1]{fontenc}    
\usepackage{hyperref}       
\usepackage{url}            
\usepackage{booktabs}       
\usepackage{amsfonts}       
\usepackage{graphicx}       
\usepackage{nicefrac}       
\usepackage{microtype}      
\usepackage{xcolor}         
\usepackage{amsmath}
\usepackage{cleveref}       
\usepackage{paralist}
\usepackage{enumitem}
\usepackage{multirow}
\usepackage{xcolor}

\title{Stable Part Diffusion 4D: Multi-View RGB and Kinematic Parts Video Generation}

\author{%
  Hao Zhang$^{1,2}$\thanks{Work done as a research intern at Stability AI. \\ Project page: \texttt{https://stablepartdiffusion4d.github.io/}} \quad
  Chun-Han Yao$^{1}$ \quad
  Simon Donn\'e$^{1}$ \quad
  Narendra Ahuja$^{2}$ \quad
  Varun Jampani$^{1}$  \\
  $^{1}$Stability AI \quad
  $^{2}$University of Illinois Urbana-Champaign \\
}

\begin{document}

\maketitle


\begin{figure}[ht]
    \centering
    \includegraphics[width=1\textwidth]{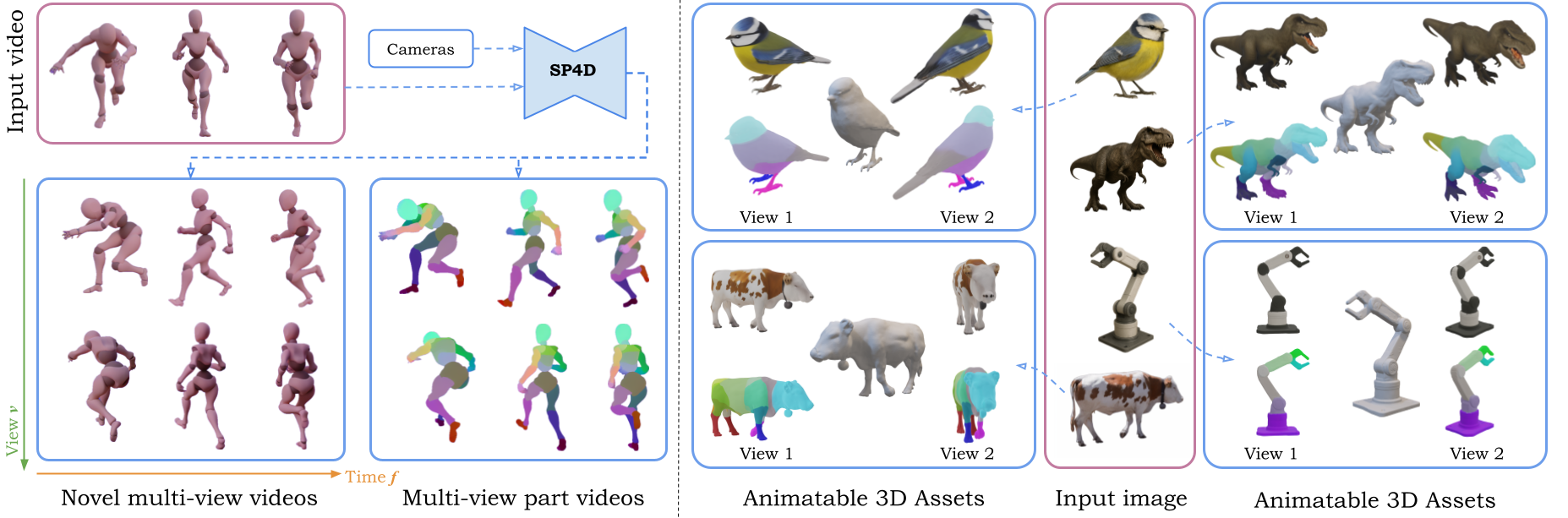}
    \vspace{-10pt}
    \caption{\textbf{Left:} \textbf{Stable Part Diffusion 4D (SP4D)}  takes a monocular input video and generates novel-view RGB videos (bottom-left) as well as consistent part segmentation videos across all views. 
    \textbf{Right:} SP4D also supports single image input and synthesizes multi-view RGB images and corresponding part decompositions. These results can be lifted to 3D to produce riggable meshes with part-aware geometry and articulated structure.  
    }
    \vspace{0pt}
    \label{fig:teaser}
\end{figure}

\begin{abstract}
We present Stable Part Diffusion 4D (SP4D), a framework for generating paired RGB and kinematic part videos from monocular inputs. Unlike conventional part segmentation methods that rely on appearance-based semantic cues, SP4D learns to produce kinematic parts --- structural components aligned with object articulation and consistent across views and time.
SP4D adopts a dual-branch diffusion model that jointly synthesizes RGB frames and corresponding part segmentation maps. To simplify architecture and flexibly enable different part counts, we introduce a spatial color encoding scheme that maps part masks to continuous RGB-like images. This encoding allows the segmentation branch to share the latent VAE from the RGB branch, while enabling part segmentation to be recovered via straightforward post-processing. A Bidirectional Diffusion Fusion (BiDiFuse) module enhances cross-branch consistency, supported by a contrastive part consistency loss to promote spatial and temporal alignment of part predictions.
We demonstrate that the generated 2D part maps can be lifted to 3D to derive skeletal structures and harmonic skinning weights with few manual adjustments. To train and evaluate SP4D, we construct KinematicParts20K, a curated dataset of over 20K rigged objects selected and processed from Objaverse XL~\citep{deitke2023objaversexl}, each paired with multi-view RGB and part video sequences. Experiments show that SP4D generalizes strongly to diverse scenarios, including real-world videos, novel generated objects, and rare articulated poses, producing kinematic-aware outputs suitable for downstream animation and motion-related tasks.
\end{abstract}

\section{Introduction}

Generating kinematic-aware and structure-consistent videos from monocular inputs is a fundamental challenge in computer vision and graphics, with wide applications in animation, AR/VR, robotics, and simulation. A key aspect of this is understanding how an object moves, articulates, and preserves spatial part relationships over time. While conventional video generation methods focus on realistic RGB synthesis, they often overlook internal articulation and fail to model a consistent structure.

\begin{figure}[t!]
    \centering
    \includegraphics[width=1\textwidth]{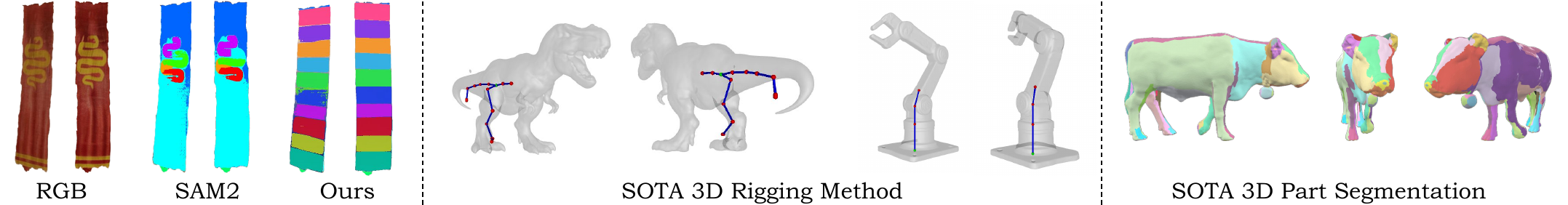}
    \vspace{-10pt}
    \caption{\textbf{Limitations of traditional 2D and 3D part decomposition methods.} 
    Left: Appearance-based 2D segmentation methods like SAM2 fail to produce kinematic parts. 
    Middle: SOTA 3D rigging methods~\citep{song2025magicarticulate} lack the capability to infer kinematic part structures from appearance and generalize poorly to diverse shapes. 
    Right: Existing 3D part segmentation models~\citep{tang2024segment,yang2024sampart3d} focus on semantic regions and are not suited for kinematic decomposition.}
    \vspace{-10pt}
    \label{fig6}
\end{figure}

Recent 4D generation approaches~\citep{liang2024diffusion4d,zhang20244diffusion,li2024vividzoo,ren2024l4gm,xie2024sv4d,yang2024diffusion,zhao2024genxd,zhu2025ar4d,yao2025sv4d} have made notable progress in reconstructing dynamic 3D sequences from monocular video, but primarily concentrate on surface-level geometry. These methods do not provide meaningful structural part decomposition and are not optimized for articulated modeling. Auto-rigging methods are a traditional option for extracting kinematic parts. However, learning-based rigging methods~\citep{xu2020rignet,liu2025riganything,song2025magicarticulate,zhang2025one,deng2025anymatedatasetbaselineslearning} operate on static 3D meshes and rely on explicit supervision such as skeletal annotations or pre-rigged models. However, these methods are fundamentally constrained by the limited scale and diversity of high-quality 3D rigging datasets, making it difficult to leverage large-scale 2D visual data and powerful pretrained image/video models. As a result, they struggle to generalize to novel object categories and rare articulated poses as shown in \cref{fig6}.

Meanwhile, part segmentation methods~\citep{amir2021deep,tang2024segment,yang2024sampart3d,yang2025holopart} often rely on semantic labels or appearance cues, leading to predictions that are temporally unstable or inconsistent across viewpoints. Most of these methods focus on semantic segmentation (e.g., head, tail, leg), which does not necessarily reflect the physical articulation or structural function of an object as shown in \cref{fig6}. In contrast, kinematic part segmentation identifies physically meaningful regions that move together over time—providing essential structure for downstream animation, motion retargeting, or deformation modeling as shown in \cref{fig:teaser}.

In this work, we introduce \textbf{Stable Part Diffusion 4D (SP4D)}, a novel framework for jointly generating RGB and kinematic part videos from monocular inputs.
SP4D builds on multi-view video diffusion~\citep{yao2025sv4d} and adopts a dual-branch architecture: one UNet generates multi-view RGB frames, while the other produces spatially and temporally consistent part segmentation maps. Unlike conventional approaches that rely on predefined semantic categories or part counts, we encode part masks as continuous RGB-like images using a spatial color encoding scheme. This allows the part branch to share the same VAE encoder and decoder with the RGB branch, and enables discrete part maps to be recovered via simple clustering in post-processing.

To ensure coherence between appearance and structure, we introduce a novel \textbf{Bidirectional Diffusion Fusion (BiDiFuse)} module, inspired by~\citet{vainer2024collaborative}, which facilitates information exchange between the RGB and part branches during the denoising process. This cross-branch communication encourages mutual guidance and alignment between modalities.
Crucially, because of the parts' spatial color encoding, the diffusion model lacks explicit supervision to enforce consistent part appearance across different views and time steps. The resulting temporal inconsistency leads to severe degradation in structural coherence. To address this, we introduce a \textbf{contrastive part consistency loss}, which aligns latent part features corresponding to the same physical regions across views and time. This loss plays a central role in enabling the model to learn stable, kinematically meaningful part representations that remain consistent throughout the generated video.

Although our framework does not explicitly output 3D models, it enables a lightweight rigging pipeline by lifting the 2D part maps to 3D. From the recovered part regions, we estimate harmonic skinning weights without requiring explicit skeleton annotations—allowing the generated videos to support animation-aware applications with minimal manual intervention.

To support training and evaluation, we curate \textbf{KinematicParts20K}, a dataset of over 20K rigged objects selected and processed from Objaverse XL~\citep{deitke2023objaversexl}, annotated with skinning weights. We adopt a two-stage training strategy: the model is first trained on ObjaverseDy~\citep{xie2024sv4d} with RGB supervision only and the BiDiFuse module bypassed; it is then fine-tuned on KinematicParts20K with supervision on both branches. This strategy leverages the generalization strength of pre-trained RGB diffusion models while gradually introducing structure-aware learning.

\vspace{0.5em}
\noindent \textbf{Our main contributions are as follows:}
\begin{itemize}[leftmargin=*]
\item We propose \textbf{Stable Part Diffusion 4D (SP4D)}, the first framework to generate multi-view, temporally consistent kinematic part decompositions jointly with RGB videos from monocular inputs.
\item We introduce a compact architecture with \textbf{spatial color encoding} for encoder-decoder sharing, efficient joint modeling with our novel \textbf{Bidirectional Diffusion Fusion (BiDiFuse)} and a \textbf{contrastive part consistency loss} to explicitly enforce cross-view and temporal alignment of part features.
\item We establish a simple yet effective \textbf{2D-to-Kinematic Mesh pipeline} by lifting part maps to 3D and estimating harmonic skinning weights, enabling skeleton-free animation-ready outputs.
\item We curate \textbf{KinematicParts20K}, a large-scale dataset of over 20K rigged objects with paired RGB and part video annotations to support training and evaluation.
\item Our method demonstrates strong generalization across real-world and synthetic scenarios, and offers a promising direction for leveraging 2D data and pretrained priors to solve long-standing challenges in 3D rigging, with clear benefits for downstream animation and motion-related tasks.
\end{itemize}

\section{Related Work.}

\textbf{3D and 4D Generation.}
We focus on diffusion-based 3D and 4D generation, which typically yield higher-quality assets than feed-forward techniques~\citep{hong2023lrm,jiang2022LEAP,wang2023pf,zou2023triplane,wei2024meshlrm,tochilkin2024triposr,ren2024l4gm,chen2024v3d,zuo2024videomv} and are not as class-bound as a GAN or VAE.
We identify three main approaches: SDS-based and photogrammetry-based methods leverage recent advantages in image and video diffusion models, compared to directly performing diffusion in 3D.
The seminal Dreamfusion~\citep{poole2022dreamfusion} used Score Distillation Sampling to refine a random initialization using the diffusion model; it is training-free but takes a very high inference cost.
Follow-up works have improved both the quality and the inference speed significantly~\citep{yi2023gaussiandreamer,tang2023dreamgaussian,shi2023mvdream,wang2024prolificdreamer,li2023sweetdreamer,weng2023consistent123,pan2023enhancing,chen2023text,sun2023dreamcraft3d,sargent2023zeronvs,EnVision2023luciddreamer,zhou2023dreampropeller,guo2023stabledreamer}, but these methods are still considered impractically slow for many contexts.
An alternative way of leveraging existing image and video models is to synthesize multi-view imagery and subsequently perform photogrammetry to extract 3D structure~\citep{liu2023zero,liu2023syncdreamer,long2023wonder3d,voleti2024sv3d,ye2023consistent,karnewar2023holofusion,instant3d2023,shi2023toss,shi2023zero123++,wang2023imagedream,liu2024one,liu2023one2}).
Finally, more recent methods perform the modeling directly in 3D~\citep{zhao2025hunyuan3d,xiang2024trellis}, often relying on powerful VAEs to compress the data dimensionality and make the problem tractable.

Image-based approaches tend to be much more data-efficient, as they can leverage the strong priors of the underlying diffusion models, at the price of costlier inference or training.
However, both families of approaches omit a key aspect for practical usability: rigging and skinning of the objects, to turn them into animation-compatible assets.
We propose to generate kinematics-aware part segmentation for the resulting objects, which feeds more cleanly into downstream pipelines.

\textbf{Rigging and animation.}
Preparing raw 3D objects for animation involves two steps: rigging (building a piecewise-rigid skeleton of bones) and skinning (defining how each part of the object deforms in function of the movement of these bones).
We focus on the former, as a bad rig precludes proper skinning.
Although we do not provide an end-to-end rigging and skinning pipeline, the kinematic parts we segment are a proxy to both: they form local clusters that directly correlate to the most influential bone; we posit that bones and even the skeletal tree structure can be extracted from these.

Learning-based rigging methods~\citep{xu2020rignet,liu2025riganything,song2025magicarticulate,zhang2025one,deng2025anymatedatasetbaselineslearning} have shown promise in predicting skeleton structure and inferring skinning weights, but are typically trained on very limited datasets, restricting their generalization to unseen object categories and poses.
Moreover, most of them operate on static geometry and fail to capture and/or leverage the dynamic articulation present in real-world videos.
Our approach addresses these limitations by leveraging the rich video diffusion priors and learning movement-aware part decomposition from videos, thereby enabling broader generalization and dynamic rigging capabilities from very little training data.

\textbf{Part Decomposition.}
Parts are useful intermediate representations for recognition, generation, and animation. Earlier 3D segmentation methods~\citep{qi2017pointnet++,li2018pointcnn,qian2022pointnext} rely on static geometry and annotated datasets, which limits their generalization to unseen or dynamic objects. More recent works use 2D semantic features for co-part segmentation~\citep{hung2019scops,amir2021deep,tang2024segment,yang2024sampart3d}, but these tend to be view-inconsistent and temporally unstable. Moreover, semantic parts are not always meaningful for animation, where rigid or articulable components are preferred.
As shown in \cref{fig6}, the semantic and kinematic parts differ both visually and functionally. Our goal is to identify physically coherent regions that move consistently over time. To our knowledge, no prior work explicitly tackles kinematic part segmentation—likely due to the lack of suitable training data. We address this by leveraging the pretrained SV4D~\citep{yao2025sv4d} video diffusion model, and extending it with a parallel part segmentation branch trained in the multi-view video space, following~\citet{vainer2024collaborative}.

\section{Method}

\subsection{Preliminaries: SV4D 2.0 Network Architecture}

Our method builds on the SV4D 2.0 framework~\citep{yao2025sv4d}, a state-of-the-art multi-view video diffusion model designed for 4D content generation. SV4D 2.0 synthesizes multi-frame, multi-view videos using a spatio-temporally consistent latent diffusion architecture. It takes either a monocular video or a single still image as input.
At its core, SV4D 2.0 represents video as a latent tensor (in 4D, as indexed by spatial, temporal, and view dimensions) and applies denoising through a UNet-based architecture composed of spatial, temporal, and view-level attention blocks.
The architecture initializes frame attention modules from Stable Video Diffusion~\citep{blattmann2023stable} and spatial-view components from SV3D~\citep{voleti2024sv3d}, benefiting from strong spatio-temporal priors. The resulting model supports long-range, self-consistent video synthesis and handles both large deformations and occlusions robustly.
It also introduces learnable $\alpha$-blending strategies to combine temporal and spatial-view features during fusion, enabling smooth integration of multiple priors while preserving the pre-trained knowledge.
The model is conditioned on both camera and frame embeddings, allowing flexible synthesis under diverse trajectories and temporal contexts. During training, SV4D 2.0 applies random view masking, which reduces reliance on explicit multi-view supervision and allows inference without external view-conditioning models.
To improve performance across sparse or nonuniform camera layouts, the model replaces traditional view attention with 3D attention layers that jointly reason over spatial and view axes. 


\subsection{Stable Part Diffusion 4D}

\begin{figure}[t!]
    \centering
    \includegraphics[width=1\textwidth]{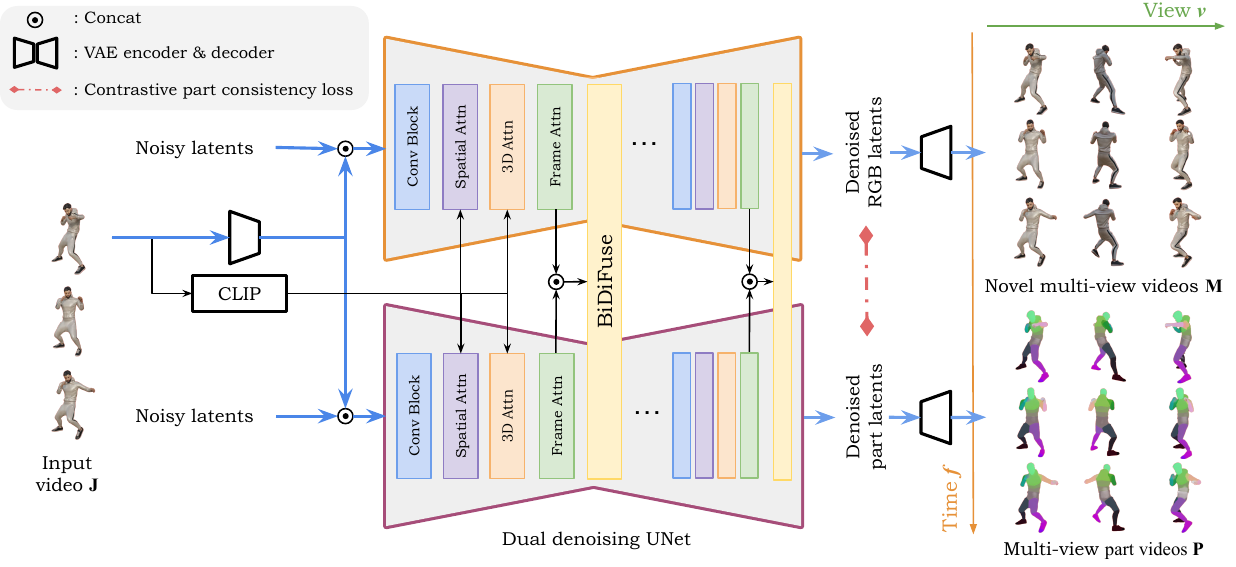}
    \vspace{-10pt}
    \caption{\textbf{Stable Part Diffusion 4D model architecture.} 
    Our model builds upon SV4D 2.0 and extends it with a parallel part segmentation branch and a BiDiFuse module that enables bidirectional feature exchange between RGB and part branches. The network jointly generates multi-view videos for appearance and kinematics-aware part segmentation. Key components include: (1) spatial color encoding for part masks, enabling shared VAE encoder/decoder; (2) BiDiFuse for cross-branch consistency; and (3) a contrastive loss for spatial-temporal part alignment. We use a two-stage training strategy: first, training the RGB branch on ObjaverseDy, then fine-tuning the full model with BiDiFuse on KinematicParts20K with supervision on both branches.}

    \label{fig:network}
\end{figure}

\textbf{Problem Setting.}
Stable Part Diffusion 4D (SP4D) aims to generate multi-view, temporally coherent kinematic part segmentation videos alongside consistent RGB videos, conditioned on a monocular RGB video input.
Formally, conditioned on input frames $J = \{J_f\}_{f=1}^F$, the model aims to produce
\[
M = \{M_{v,f}\}_{v=1,f=1}^{V,F}, \quad P = \{P_{v,f}\}_{v=1,f=1}^{V,F},
\]
where $M_{v,f}$ and $P_{v,f}$ represent the generated RGB image and its corresponding part segmentation at view $v$ and frame $f$, respectively.

The goal is to produce photo-realistic video sequences $M$ that are consistent across views and time, while also generating part segmentations $P$ that reflect view-invariant kinematic structure. Unlike traditional semantic segmentation, these part representations are not predefined by category, but instead capture locally rigid, articulated motion-consistent regions.

\textbf{Network Architecture.}
Our network architecture extends SV4D 2.0 with a dual-branch UNet structure to jointly generate RGB and kinematic part segmentation videos. We adopt the full SV4D 2.0 backbone for RGB generation, including its spatial, temporal, and 3D attention mechanisms, and replicate this backbone to form a second branch for part segmentation, as seen in \cref{fig:network}.

Each branch processes half of the latent channels and shares positional embeddings (e.g., camera intrinsics and temporal indices) as input. Inspired by~\cite{vainer2024collaborative}, the two branches operate independently but are connected through dedicated Bidirectional Diffusion Fusion (BiDiFuse) layers inserted at every block of the network. 
Given intermediate features $h^{\text{RGB}}$ and $h^{\text{Part}}$ at any resolution, we compute updated representations using a fusion module:
\begin{equation}
    h^{\text{RGB}}_{\text{fused}} = h^{\text{RGB}} + \mathcal{F}\left([h^{\text{RGB}}, h^{\text{Part}}]\right),
    \quad
    h^{\text{Part}}_{\text{fused}} = h^{\text{Part}} + \mathcal{F}\left([h^{\text{RGB}}, h^{\text{Part}}]\right)
\label{eq:fused}
\end{equation}
where $\mathcal{F}
$ is a lightweight fusion function composed of two $1\times1$ convolutions with ReLU activations. This module encourages bidirectional feature sharing while maintaining branch-specific learning.

The forward pass proceeds as follows: the input latent is split along the channel dimension and passed through two identical UNet backbones. After each encoder block, the intermediate features are fused via BiDiFuse. The same process applies to the middle block and each decoder stage, with skip connections preserved within corresponding branches. The final outputs from both branches are separately passed through a shared VAE decoder to produce RGB and part predictions independently.

\textbf{Spatial Color Encoding.} 
To enable decoder sharing between RGB and part branches, we represent part segmentation maps as continuous RGB-like images using a spatial color encoding scheme. To assign temporally consistent colors, we first normalize the 3D coordinates of each point on the mesh or reconstructed surface to a unit cube. Then we compute the coordinates of the 3D center of each part in the first frame and use its normalized $(x, y, z)$ coordinate as the color code for all frames and views. This ensures that the same part is assigned the same color across all frames and views, maintaining identity consistency over time. Unlike schemes that randomly assign colors to parts per iteration, our deterministic encoding significantly reduces computational overhead, as random coloring would require regenerating encoded part images at every step. Our approach enables the diffusion model to treat part segmentation as an image generation task, facilitating compatibility with the RGB branch and enabling unified training within a shared latent space.

\textbf{Back-Mapping from RGB Image to Part Mask.} 
To recover discrete part masks from the generated spatial encoding, we avoid clustering the generated colors, which can be noisy. Instead, we apply SAM (Segment Anything Model) in auto-generation mode to produce per-view segmentation masks --- we found this remarkably effective at providing clean candidate segments. For each segment, we then compute the mode of the underlying RGB pixel values and assign this color to the entire mask. This procedure robustly eliminates pixel-level noise and ensures clean, discrete part representations. Then we apply clustering~\citep{mcinnes2017hdbscan} on all images to produce part masks. We do not use SAM2 (the video tracking version of SAM) as it only supports parts that are visible in the first frame, and thus fails to capture parts that only first appear later in the video.

\textbf{Contrastive Part Consistency loss.}
The spatial color encoding represents parts as RGB-like images, enabling a shared encoder and decoder between both branches. However, this model lacks an explicit supervision to ensure that the same kinematic part maintains a consistent appearance across different viewpoints and time steps. Without regularization, the model may produce temporally or spatially inconsistent segmentations.
To address this, we extract part-specific features by aggregating pixel-level features within each predicted part region, and project them into a shared embedding space. For each training batch, we collect a set of part features $\{f_i\}_{i=1}^N$, where each $f_i$ corresponds to one part instance (across view and frame). Features with the same part identity but from different frames or views are considered positive pairs, while features from different parts serve as negatives.
We adopt an InfoNCE-style contrastive loss defined over all part pairs~\citep{oord2018infonce}:
\begin{equation}
\mathcal{L}_{\text{contrast}} =
- \mathbb{E}_{i \in \mathcal{P},\, j \in \mathcal{P}_i^+} 
\left[
    \log \frac{\exp(\text{sim}(f_i, f_j)/\tau)}
    {\sum_{k \in \mathcal{P} \setminus \{i\}} \exp(\text{sim}(f_i, f_k)/\tau)}
\right]
\label{eq:contrast}
\end{equation}
where $\mathcal{P}$ is the set of all valid part features, $\mathcal{P}_i^+$ is the set of positive indices for part $i$, $\text{sim}(\cdot,\cdot)$ denotes cosine similarity, and $\tau$ is a temperature hyperparameter ($\tau = 0.07$ by default). This loss encourages the same part to be consistently encoded across views and frames, while remaining distinct from different parts.

\subsection{KinematicParts20K Dataset}

We curate the KinematicParts20K dataset from the ObjaverseDy++~\citep{yao2025sv4d} dataset to support the training and evaluation of our part-aware generation framework. We first filter objects that include rigging annotations with well-structured skeletal hierarchies and bone transformations.

\textbf{Bone Merging to Control Granularity.} For objects with excessively many bones, we automatically reduce skeletal complexity. For each pair of connected bones, we compute (1) the average relative 3D displacement between the two bones across all frames, and (2) the cosine similarity between their 2D part segmentation masks based on Dino features. If both the motion difference and feature dissimilarity fall below predefined thresholds, we merge the two bones. We set an upper bound of 100 bones per object; if an object cannot be reduced to within this limit, it is discarded from the dataset.

\textbf{Multiview Rendering and Part Label Generation.} For each selected object, we render 24 frames from 24 camera views uniformly distributed along a horizontal circle. We also render per-bone 2D skinning weight maps. To compute the part segmentation masks, we use a per-pixel argmax over all bone-specific weight maps within each view. The resulting part maps provide high-quality multiview kinematic part segmentation labels aligned with the rigging annotations, enabling supervised training of SP4D using part labels that reflect true kinematic decomposition.
After all the filtering steps, we are left with almost 20,000 training objects.

\subsection{2D-to-Kinematic Mesh Generation}

\textbf{Lifting from 2D.}
We propose a simple yet effective pipeline that converts a single image into a fully riggable 3D asset with geometry, part decomposition, and skinning weights -- only missing skeletal connectivity.
We first apply our Stable Part Diffusion 4D (SP4D) model to generate multi-view sequences of RGB frames and corresponding part segmentation (cleaned by SAM) from a single input image.
For recovering geometry, we use Hunyuan 3D 2.0~\citep{zhao2025hunyuan3d}, a state-of-the-art images-to-3D framework, to turn the multi-view RGB images generated by SP4D into untextured 3D geometry. Once we obtain the 3D mesh, we reuse this geometry to associate texture information from both the RGB and part segmentation views separately, following Hunyuan 3D 2.0. Through HDBSCAN~\citep{mcinnes2017hdbscan}, we assign each vertex its discrete ID for the part segmentation.

\textbf{Harmonic Skinning Weight Computation.}
Given the 3D part labels, we compute continuous skinning weights using harmonic field estimation.
We extract the boundary $\partial \Omega_p$ of part $p$ by identifying mesh edges that connect two vertices belonging to different parts; the binary indicator function $b_p(x)$ indicates whether vertex $x$ belongs to part $p$. We then solve:
\begin{align*}
\Delta w_p(x) = 0 \quad \text{for all interior vertices,} \quad \text{subject to} \quad w_p(x) = b_p(x) \text{ on } \partial \Omega_p
\end{align*}
where $\Delta$ is the mesh Laplacian operator and $w_p(x)$ denotes the smooth harmonic field corresponding to part $p$. The harmonic solution to this Laplace equation propagates part influence across the surface, yielding soft per-vertex part assignments which we interpret as skinning weights.

\section{Experiments}

\begin{figure}[ht]
    \centering
    \includegraphics[width=.95\textwidth]{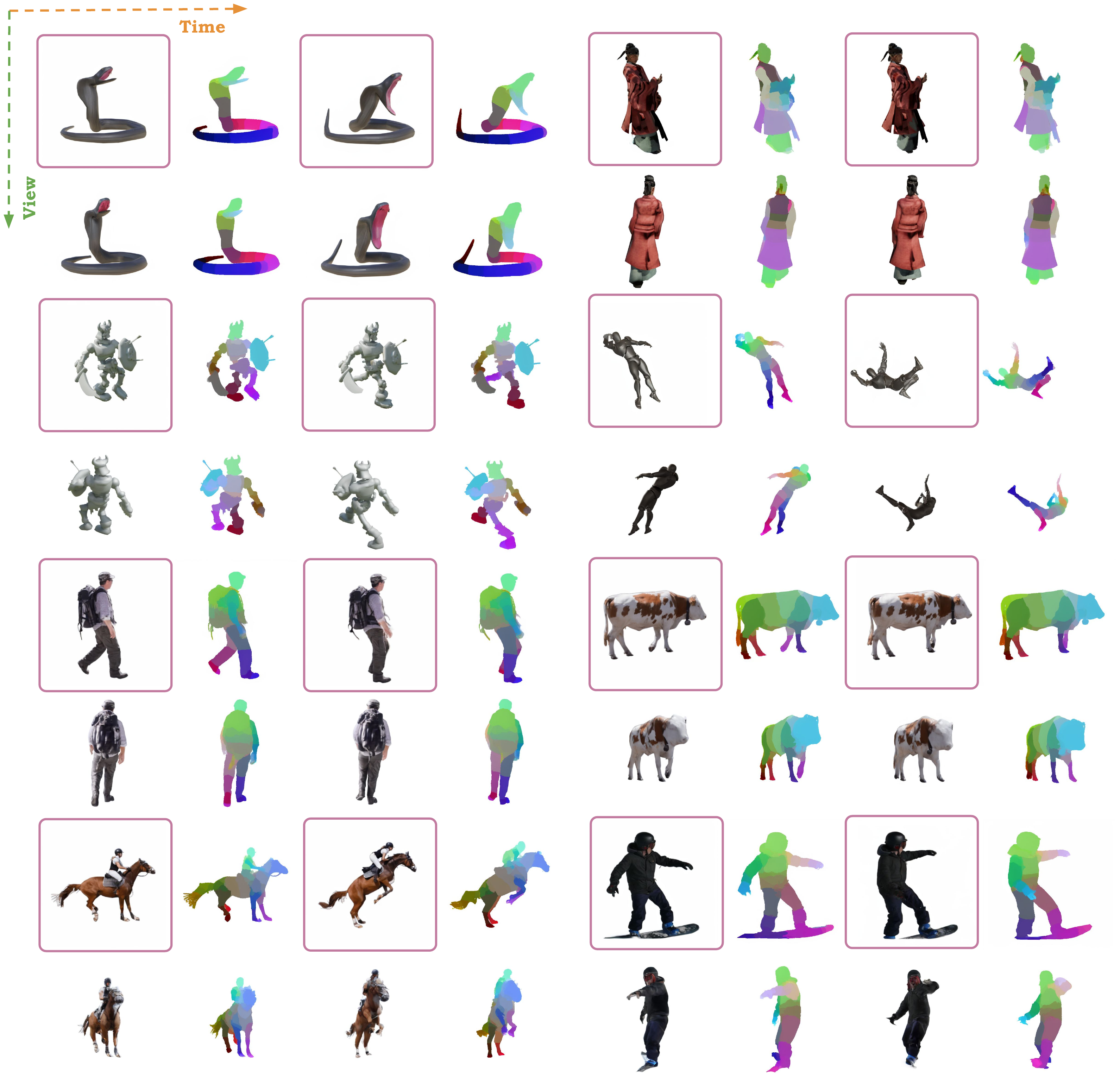}
    \vspace{-5pt}
    \caption{\textbf{Multi-view kinematic part video results on synthetic and real-world videos.} 
    We show qualitative results of our SP4D model on both the validation set of KinematicParts20K and real-world DAVIS videos. Each group presents two time frames across two novel views. The input video frame is noted with \textcolor{purple}{purple} boxes. SP4D produces temporally and spatially consistent part decompositions across diverse object categories and motions.}
    \vspace{-5pt}
    \label{fig4}
\end{figure}

We demonstrate that SP4D performs robustly and generalizes across a wide variety of articulated objects with diverse shapes and motions, including both synthetic models and real-world videos in \cref{fig4}.
We conduct comprehensive experiments to evaluate the effectiveness of our method, including comparisons with state-of-the-art approaches for part segmentation, as well as ablation studies on key design choices. We report both quantitative metrics (mIoU, ARI, F1 Score, mAcc) in~\cref{tab:multi_view_multi_frame} and a user study in~\cref{tab:userstudy} to assess quality from a rigging perspective. Details on implementation, datasets, training regime, evaluation protocols, metric definitions and more experiments on 3D segmentation and rigging can be found in the Appendix.

\subsection{Part Decomposition Comparison}

\textbf{2D Part Decomposition.}
We compare SP4D with two representative 2D part segmentation baselines. The first is SAM2~\citep{ravi2024sam}, a tracking-based method that generates part masks in the first image and propagates them to the others. We also include a stronger variant, \textbf{SAM2*}, where point prompts from the ground-truth part centroids are used to initialize tracking. 
The second baseline is DeepViT~\citep{amir2021deep}, an unsupervised segmentation method that leverages features from a self-supervised DINO-ViT model~\citep{caron2021emerging}. We apply K-Means clustering on intermediate feature maps to obtain part-level masks across views. 

\begin{figure}[t!]
    \centering
    \includegraphics[width=1\textwidth]{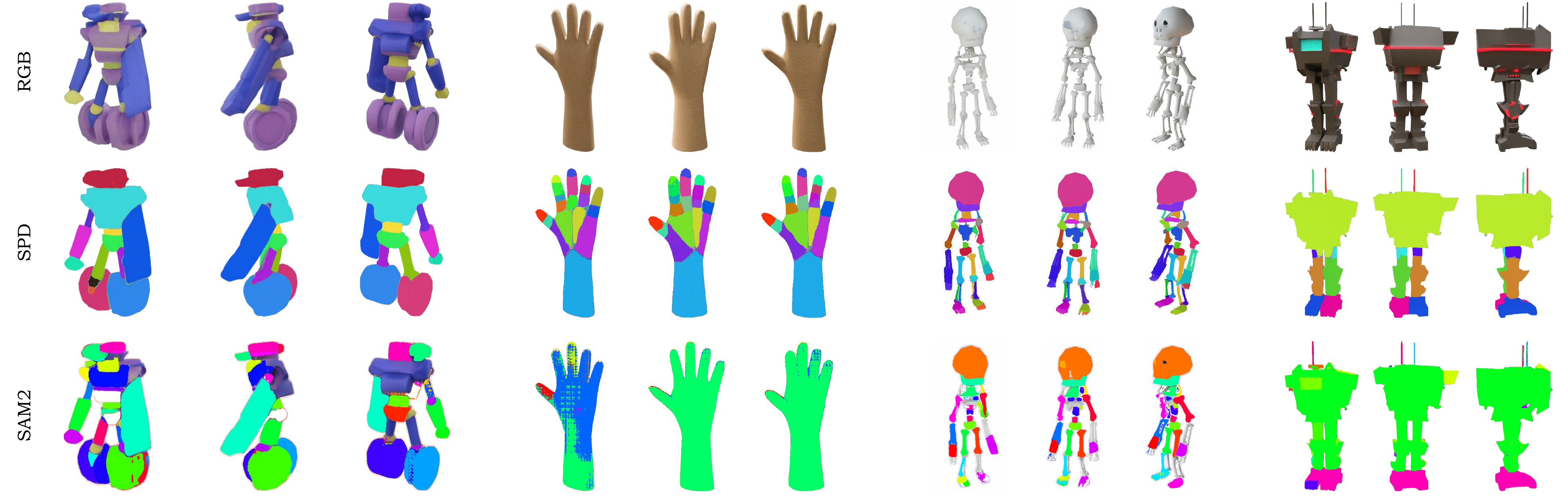}
    \vspace{-10pt}
    \caption{\textbf{Visual comparison of part segmentation.} 
    We show results across three views for various articulated objects. The rows contain input RGB image (top), our SP4D-generated part segmentation (middle), and the SAM2 baseline (bottom). Compared to SAM2, SP4D produces more structured part decompositions that align with object articulation and are consistent across views.}
    \label{fig5}
\end{figure}

As shown in \cref{tab:multi_view_multi_frame}, SP4D significantly outperforms all baselines in both multi-view and multi-frame settings. \textbf{SAM2} performs poorly due to its dependency on appearance and semantic cues, which rarely align with kinematic part boundaries.
\textbf{SAM2*}, despite being guided by ground-truth points, suffers from the same fundamental limitation. While DeepViT captures coarse semantic structures, it lacks any awareness of object articulation or motion consistency.
In contrast, SP4D generates parts directly via kinematic-aware diffusion, leveraging geometry and view consistency to produce temporally stable, kinematic-aware decompositions. This advantage is further supported by \cref{fig5}, which presents qualitative comparisons across multiple views, clearly showing SP4D’s superior structural alignment. Additional user preference results in \cref{tab:userstudy} also confirm the perceptual quality of SP4D’s outputs.

\begin{table}[t]
\centering
\begin{minipage}[t]{0.63\textwidth}
\centering
\caption{
\textbf{Quantitative comparison of kinematic parts} on KinematicParts20K val set for \textbf{multi-view} (static object) and \textbf{multi-frame} (static camera).
The Hungarian algorithm aligns predictions to the ground-truth, ignoring parts missing in the first image. 
\textbf{SAM2*} uses ground-truth point prompts per part.
}
\label{tab:multi_view_multi_frame}
\resizebox{\textwidth}{!}{%
\begin{tabular}{lcccccccc}
\toprule
\multirow{2}{*}{\textbf{Method}} & \multicolumn{4}{c}{\textbf{Multi-view}} & \multicolumn{4}{c}{\textbf{Multi-frame}} \\
& mIoU & ARI & F1 & mAcc & mIoU & ARI & F1 & mAcc \\
\midrule
SAM2                       & 0.15 & 0.05 & 0.31 & 0.21 & 0.16 & 0.05 & 0.32 & 0.22 \\
SAM2*                      & 0.22 & 0.08 & 0.37 & 0.26 & 0.34 & 0.16 & 0.45 & 0.34 \\
DeepViT                    & 0.17 & 0.06 & 0.33 & 0.23 & 0.18 & 0.06 & 0.34 & 0.24 \\
Ours w/o PCP Loss          & 0.38 & 0.15 & 0.46 & 0.49 & 0.44 & 0.22 & 0.52 & 0.56 \\
Ours w/o BiDiFuse          & 0.57 & 0.51 & 0.60 & 0.62 & 0.61 & 0.58 & 0.64 & 0.68 \\
\textbf{Ours (Full)}       & \textbf{0.68} & \textbf{0.60} & \textbf{0.70} & \textbf{0.74} & \textbf{0.70} & \textbf{0.63} & \textbf{0.72} & \textbf{0.77} \\
\bottomrule
\end{tabular}}
\end{minipage}
\hfill
\begin{minipage}[t]{0.35\textwidth}
\centering
\caption{\textbf{User study on kinematic part segmentation.} 
Participants rated three methods on part clarity, view consistency, and rigging suitability. 
The study was conducted on 20 randomly selected samples from the validation set.}
\label{tab:userstudy}
\resizebox{\textwidth}{!}{%
\begin{tabular}{lccc}
\toprule
\textbf{Method} & \textbf{Ours} & SAM2 & DeepViT \\
\midrule
Clarity & \textbf{4.42} & 2.13 & 2.01 \\
Consistency & \textbf{4.09} & 2.00 & 1.86 \\
Rigging & \textbf{4.26} & 1.75 & 1.69 \\
Average & \textbf{4.26} & 1.96 & 1.85 \\
\bottomrule
\end{tabular}}
\end{minipage}
\end{table}

\textbf{3D Part Decomposition.}
Recent state-of-the-art 3D part segmentation methods, such as SAMesh~\citep{tang2024segment} and SamPart3D~\citep{yang2024sampart3d}, rely on 2D segmentation cues to supervise 3D decomposition in different ways. SAMesh fuses 2D segmentations (from SAM) of multiple rendered views using visibility-weighted voting. In contrast, SamPart3D distills dense visual features from DINOv2 into a 3D point-based backbone, and leverages SAM masks through a scale-conditioned MLP to achieve granularity-controllable part grouping via clustering.

Despite their differences, both methods fundamentally depend on the quality of 2D segmentations. When appearance-based segmenters like SAM or DINOv2 fail to produce meaningful part boundaries—particularly for kinematic or textureless regions—the resulting 3D decomposition is unreliable and misaligned with object articulation. As illustrated in \cref{fig6}, these approaches often struggle to produce structurally coherent part segmentation under such challenging conditions. Additional comparisons are provided in the supplementary material.

\subsection{Ablation Study}

We conduct ablation studies to evaluate the contribution of two core components in our framework: the BiDiFuse cross-branch fusion module and the part consistency loss. Results are reported under both multi-view and multi-frame settings (see \cref{tab:multi_view_multi_frame}).
Removing the part consistency loss leads to noticeable performance degradation, especially in terms of ARI. Without this loss, the model loses explicit guidance to maintain spatial and temporal coherence of part assignments across views or frames, resulting in fragmented or inconsistent segmentations. This highlights the importance of encouraging feature-level alignment among corresponding parts throughout the video sequence.

Disabling the BiDiFuse module also causes substantial drops across all metrics. Since BiDiFuse facilitates bidirectional interaction between the RGB and part branches.
Without it, the network lacks effective cross-modal information exchange, leading to suboptimal alignment between both branches, particularly in view-consistency and part boundary sharpness.
Crucially, the segmentation branch can no longer effectively leverage the prior of the RGB model.
These results confirm that both components are essential for achieving robust, consistent, and rigging-friendly part decompositions.

\section{Broader Societal Impact}
%

SP4D has the potential to substantially reduce the manual work required for rigging in animation and 3D asset production, benefiting creators in film, gaming, education, AR/VR, and robotics.
Particularly for those with limited access to professional modeling pipelines, our method broadens accessibility to animation-ready assets and opens new opportunities in educational content creation, interactive media, and rapid prototyping.
SP4D’s ability to generalize across real-world footage and synthetic objects further supports its potential in democratizing digital content creation.

However, the ability to synthesize from minimal visual input introduces risks. These include the creation of synthetic humans or avatars for deceptive purposes. While our method does not focus on facial reenactment or human identity synthesis, downstream misuse remains a concern. We recommend clear disclosure, attribution mechanisms for automatically generated 3D content, and ethical oversight in such applications.
We emphasize that all training uses CC-licensed assets, carefully filtered to respect creator rights.

\section{Conclusion}

We propose SP4D to jointly generate multi-view parts video with aligned RGB frames from a monocular input video.
Uniquely, we predict kinematic rather than semantic parts, based on segments in articulated motion skeletons.
This closes a significant gap in the 3D generation pipeline, drastically reducing the manual annotation required to prepare the generated objects for animation.

By leveraging a synchronized two-branch architecture, we maximally leverage the prior of the pre-trained RGB model; this results in a robust and generalizable approach despite the data scarcity for training.
Both a quantitative comparison with representative baselines and a user study show the clear benefit of our approach over existing semantic-oriented part segmentation for the same task.

{
    \small
    \bibliographystyle{ieeenat_fullname}
    \bibliography{neurips_2025}
}

\clearpage

\newpage
\section*{NeurIPS Paper Checklist}

\begin{enumerate}

\item {\bf Claims}
    \item[] Question: Do the main claims made in the abstract and introduction accurately reflect the paper's contributions and scope?
    \item[] Answer: \answerYes{}.
    \item[] Justification: {Any significant claim made in our manuscript is at least outlined in the abstract and motivated in the introduction. We carefully note that we do not address the whole gap towards automatic rigging and skinning, but solve a significant piece of that puzzle.}

\item {\bf Limitations}
    \item[] Question: Does the paper discuss the limitations of the work performed by the authors?
    \item[] Answer: \answerYes{} 
    \item[] Justification: {We include a limitation section in the paper.}

\item {\bf Theory assumptions and proofs}
    \item[] Question: For each theoretical result, does the paper provide the full set of assumptions and a complete (and correct) proof?
    \item[] Answer: \answerNA{} 
    \item[] Justification: {We offer no purely theoretical results.}

    \item {\bf Experimental result reproducibility}
    \item[] Question: Does the paper fully disclose all the information needed to reproduce the main experimental results of the paper to the extent that it affects the main claims and/or conclusions of the paper (regardless of whether the code and data are provided or not)?
    \item[] Answer: \answerYes{} 
    \item[] Justification: {We have included the pertinent details around model architecture, data sourcing and curation, and evaluation approach. Due to space constraints, most of these are deferred to the Appendix.}

\item {\bf Open access to data and code}
    \item[] Question: Does the paper provide open access to the data and code, with sufficient instructions to faithfully reproduce the main experimental results, as described in supplemental material?
    \item[] Answer: \answerNo{} 
    \item[] Justification: {Unfortunately, we cannot commit to releasing the code at this time.}

\item {\bf Experimental setting/details}
    \item[] Question: Does the paper specify all the training and test details (e.g., data splits, hyperparameters, how they were chosen, type of optimizer, etc.) necessary to understand the results?
    \item[] Answer: \answerYes{} 
    \item[] Justification: {We include these details in the main paper and appendix.}

\item {\bf Experiment statistical significance}
    \item[] Question: Does the paper report error bars suitably and correctly defined or other appropriate information about the statistical significance of the experiments?
    \item[] Answer: \answerNo{} 
    \item[] Justification: {For the quantitative results, such evaluation would be quite costly. The user study, for cost saving reasons, had only a limited number of participants and therefore also did not get error bars.}

\item {\bf Experiments compute resources}
    \item[] Question: For each experiment, does the paper provide sufficient information on the computer resources (type of compute workers, memory, time of execution) needed to reproduce the experiments?
    \item[] Answer: \answerYes{} 
    \item[] Justification: {This paper provides sufficient information on the computer resource to reproduce the experiments in the appendix.}
    
\item {\bf Code of ethics}
    \item[] Question: Does the research conducted in the paper conform, in every respect, with the NeurIPS Code of Ethics \url{https://neurips.cc/public/EthicsGuidelines}?
    \item[] Answer: \answerYes{} 
    \item[] Justification: {We have read the NeurIPS Code of Ethics and believe to be in accordance.}

\item {\bf Broader impacts}
    \item[] Question: Does the paper discuss both potential positive societal impacts and negative societal impacts of the work performed?
    \item[] Answer: \answerYes{} 
    \item[] Justification: {The main manuscript includes a discussion on the broader impacts.}
    
\item {\bf Safeguards}
    \item[] Question: Does the paper describe safeguards that have been put in place for responsible release of data or models that have a high risk for misuse (e.g., pretrained language models, image generators, or scraped datasets)?
    \item[] Answer: \answerNo{} 
    \item[] Justification: {We discuss such safeguards in the broader impacts section, but do not offer an implementation or detailed description.}

\item {\bf Licenses for existing assets}
    \item[] Question: Are the creators or original owners of assets (e.g., code, data, models), used in the paper, properly credited and are the license and terms of use explicitly mentioned and properly respected?
    \item[] Answer: \answerYes{} 
    \item[] Justification: {We have properly cited all used assets and models throughout our manuscript; please refer to those references for licensing details.}

\item {\bf New assets}
    \item[] Question: Are new assets introduced in the paper well documented and is the documentation provided alongside the assets?
    \item[] Answer: \answerNA{} 
    \item[] Justification: {We provide no new assets.}

\item {\bf Crowdsourcing and research with human subjects}
    \item[] Question: For crowdsourcing experiments and research with human subjects, does the paper include the full text of instructions given to participants and screenshots, if applicable, as well as details about compensation (if any)? 
    \item[] Answer: \answerYes{} 
    \item[] Justification: {While not present in the main manuscript, we will include these details in the appendix.}

\item {\bf Institutional review board (IRB) approvals or equivalent for research with human subjects}
    \item[] Question: Does the paper describe potential risks incurred by study participants, whether such risks were disclosed to the subjects, and whether Institutional Review Board (IRB) approvals (or an equivalent approval/review based on the requirements of your country or institution) were obtained?
    \item[] Answer: \answerNA{} 
    \item[] Justification: {There were no risks to the study participants --- we have only performed a qualitative performance assessment user study.}

\item {\bf Declaration of LLM usage}
    \item[] Question: Does the paper describe the usage of LLMs if it is an important, original, or non-standard component of the core methods in this research? Note that if the LLM is used only for writing, editing, or formatting purposes and does not impact the core methodology, scientific rigorousness, or originality of the research, declaration is not required.
    \item[] Answer: \answerNA{} 
    \item[] Justification: {LLMs were not involved in the core method development of our research.}

\end{enumerate}

\clearpage
\appendix

\begin{center}
    
{\Large \bf Stable Part Diffusion 4D:}

{\Large \bf Multi-View RGB and Kinematic Parts Video Generation}

{\Large \centering Supplementary Material}

\end{center}

In the appendix, we provide the following supplementary materials: (1) Implementation Details, (2) our newly introduced dataset, KinematicParts20K, (3) Additional Qualitative Results, and (4) Evaluation Details.

\section{Implementation Details}

Our model is implemented by directly extending the SV4D 2.0 framework~\citep{yao2025sv4d}. We retain the original U-Net architecture, latent VAE encoding, and diffusion setup, and introduce two key modifications: (1) an architecturally identical second branch that generates part segmentation outputs jointly with the existing RGB branch, and (2) Bidirectional Diffusion Fusion (BiDiFuse) modules inserted between each corresponding layer pair to enable cross-branch feature sharing. In the first stage, the RGB branch is trained following SV4D 2.0. The training setup --- including optimizer, noise schedule, loss functions, and sampling strategy --- follows SV4D 2.0 exactly. We adopt the EDM~\citep{karras2022elucidating} training framework with an L2 loss and precompute VAE latents and CLIP features for all training images to accelerate convergence. The obtained network parameters are used to initialize both the RGB and part generation branches.

We train the full SP4D model with BiDiFuse and our proposed contrastive part consistency loss on the KinematicParts20K dataset (as discussed below) for 40K iterations. Training is performed on 32 NVIDIA H100 GPUs with an effective batch size of 32, using 12 views and 4 frames per object sampled from the rendered dataset.

\section{KinematicParts20K Dataset}

Our dataset is constructed by further filtering the SV4D 2.0 dataset, which is based on CC-licensed dynamic 3D assets from Objaverse and ObjaverseXL. We select only objects that contain rigging annotations, including bone hierarchies and skinning weights. To mitigate overly fine-grained or noisy bone structures, we apply a bone merging procedure based on two criteria: (1) the relative transformation between connected bones across all frames, and (2) the similarity of their projected part appearance in 2D using DINO features. Bone pairs with low motion discrepancy and high appearance similarity are merged. Objects with more than 100 bones after merging are discarded.

All objects are scaled to unit bounding boxes and rendered at $576$\texttimes$576$ resolution using Blender's Cycles renderer under a curated set of HDRI environment maps. We adopt orbit rendering with 24 azimuthal views and 24 video frames per object. In addition to RGB, we simultaneously render per-bone skinning weight maps. For each view and frame, we generate pixel-wise part segmentation labels by taking the argmax over the bone-specific skinning maps, resulting in multi-view, multi-frame kinematic part masks for supervision.

\section{More Qualitative Results}
We show fixed-view cross-frame part tracking, fixed-frame cross-view part tracking, 3D decomposition, rigging, and animation results for synthetic data, real-world data, and zero-shot generated data.
Please refer to the summary video in the supplementary material.

\paragraph{Evaluation on 3D Segmentation.} It is also important to position SP4D in the broader context of 3D kinematic segmentation rather than only comparing against 2D segmentation baselines. 
State-of-the-art 3D segmentation methods, such as Segment Anything Mesh~\cite{tang2024segment} and SAMPart3D~\cite{yang2024sampart3d}, are built upon 2D semantic segmentation backbones (e.g., SAM, DINOv2) that are primarily texture- or appearance-driven, and thus not explicitly designed for kinematic reasoning. 
This limitation is evident in our visual comparisons (Figure.2), where appearance-based cues alone fail to recover accurate part structures for novel or textureless objects.

To quantitatively assess this gap, we conduct a comprehensive evaluation on the KinematicParts20K test set using these SOTA 3D segmentation baselines~\cite{tang2024segment,yang2024sampart3d}. 
We report mean Intersection-over-Union (mIoU), Adjusted Rand Index (ARI), F1 score, mean Accuracy (mAcc), and User Study ratings (following the evaluation criteria in Supplementary Section~D.2). 
As shown in Table~\ref{tab:segmentation}, SP4D substantially outperforms the baselines across all metrics, highlighting its capability to capture kinematic structure rather than relying solely on appearance cues.

\begin{table}[h]
\centering
\caption{Comparison of SP4D with SOTA 3D segmentation methods on KinematicParts20K-test. SP4D achieves significantly higher scores across all metrics, indicating stronger kinematic reasoning capabilities.}
\label{tab:segmentation}
\resizebox{0.6\textwidth}{!}{
\begin{tabular}{lccccc}
\toprule
\textbf{Method} & \textbf{mIoU} & \textbf{ARI} & \textbf{F1} & \textbf{mAcc} & \textbf{User Study} \\
\midrule
Segment Any Mesh & 0.15 & 0.06 & 0.29 & 0.20 & 1.98 \\
SAMPart3D        & 0.13 & 0.05 & 0.27 & 0.18 & 1.75 \\
\textbf{Ours (Full)} & \textbf{0.64} & \textbf{0.58} & \textbf{0.67} & \textbf{0.72} & \textbf{4.13} \\
\bottomrule
\end{tabular}}
\end{table}

\paragraph{Evaluation beyond segmentation accuracy.}
To further assess the usefulness of our kinematic representation beyond segmentation accuracy, we conduct additional experiments on \textit{rigging precision} and \textit{animation plausibility}.
(i) \textbf{Rigging precision.}
We evaluate the predicted skinning weights on the KinematicParts20K-test split, which contains ground-truth rigging annotations. We compare SP4D against two state-of-the-art auto-rigging methods~\cite{song2025magicarticulate,zhang2025one}, reporting precision scores in Table~\ref{tab:autorig}. SP4D achieves the highest precision ($72.7$), outperforming Magic Articulate ($63.7$) and UniRig ($64.3$), demonstrating the accuracy of our learned kinematic decomposition when ground-truth supervision is available.
(ii) \textbf{Animation plausibility for generated objects.}
For generated meshes (e.g., dinosaurs, robotic arms) without ground-truth rigging, we conduct a user study to evaluate animation plausibility. Participants were shown animations produced by SP4D and the SOTA baselines~\cite{song2025magicarticulate,zhang2025one}, and asked to rate the plausibility on a 1–5 Likert scale. SP4D achieves a significantly higher score ($4.1$) than Magic Articulate ($2.7$) and UniRig ($2.3$), confirming better generalization to unseen object categories and poses.

Notably, as shown in Figure~2 (middle), Magic Articulate, despite being trained on large-scale rigged meshes from Articulation-XL, performs well on seen categories but struggles with unusual generated shapes. In contrast, SP4D leverages strong priors from a 2D diffusion model and learns kinematic decomposition robustly, enabling accurate rigging for both real-world and synthetic objects. This highlights a key motivation for our approach: learning kinematic structure from 2D multi-view supervision yields superior generalization to novel inputs.

\begin{table}[h]
\centering
\caption{Comparison of SP4D with SOTA Auto-rigging Methods.}
\label{tab:autorig}
\resizebox{0.6\textwidth}{!}{
\begin{tabular}{lccc}
\toprule
\multirow{2}{*}{\textbf{Method}} & \multicolumn{2}{c}{\textbf{KinematicPart20K-test}} & \textbf{Generated Objects} \\
\cmidrule(lr){2-3} \cmidrule(lr){4-4}
& \textbf{Precision} & \textbf{User Study} & \textbf{User Study} \\
\midrule
Magic Articulate & 63.7 & 3.8 & 2.7 \\
UniRig           & 64.3 & 3.9 & 2.3 \\
\textbf{Ours (Full)} & \textbf{72.7} & \textbf{4.3} & \textbf{4.1} \\
\bottomrule
\end{tabular}}
\end{table}

\begin{figure}[t!]
    \centering
    \includegraphics[width=1\textwidth]{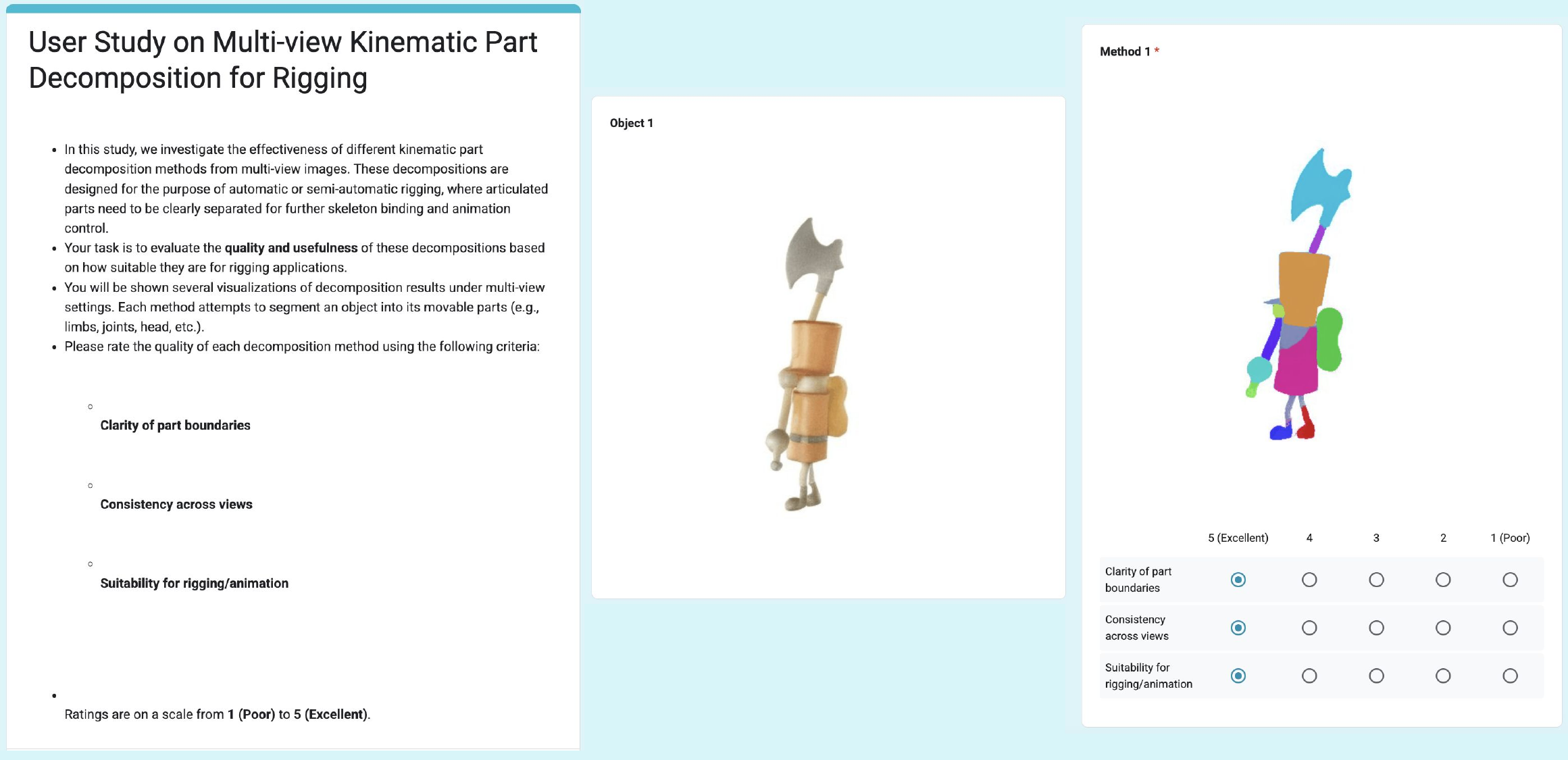}
    \vspace{0pt}
    \caption{\textbf{User study interface for evaluating multi-view kinematic part segmentation.} 
    Participants are presented with video results generated by different methods and asked to rank them based on part consistency, structural correctness, and motion coherence. The study compares SP4D with baseline methods to assess perceptual quality and kinematic alignment.}

    \vspace{0pt}
    \label{fig7}
\end{figure}

\section{Evaluation Details}

\subsection{Quantitative Metrics.}
To evaluate the quality of kinematic part decomposition across multi-view and multi-frame settings, we report four standard metrics. Since the predicted part masks are label-free, we apply the Hungarian algorithm to align predicted and ground-truth parts based on respective IoU, for those metrics which require correspondences. The following metrics are computed:

\begin{itemize}
    \item \textbf{mIoU} – Mean intersection-over-union across matched part masks.
    \item \textbf{ARI} – Adjusted Rand Index, which captures clustering similarity independent of label permutation.
    \item \textbf{F1 Score} – The harmonic mean of precision and recall, reflecting pixel-level agreement.
    \item \textbf{mAcc} – Mean class-wise accuracy, indicating the average recall per ground-truth part.
\end{itemize}

\subsection{User Study on Multi-view Kinematic Part Decomposition for Rigging}

To evaluate the practical utility of different multi-view kinematic part decomposition methods for rigging tasks, we conducted a user study focusing on the perceived quality of part segmentation from a rigging perspective. These decompositions aim to separate articulated object parts (e.g., limbs, joints, head) to facilitate automatic or semi-automatic skeleton binding and animation control.

\paragraph{Study Setup.}
We randomly selected 20 sets of decomposition results, each containing visualizations from different methods applied to the same object. For each set, we generated animated GIFs showing the part decomposition from multiple viewpoints, allowing participants to better understand spatial consistency and articulation structure. All visualizations were presented anonymously to avoid bias. The study was conducted via a Google Form and received responses from 20 participants.

\paragraph{Evaluation Criteria.}
Participants were instructed to rate each method based on the following three criteria:
\begin{itemize}
    \item \textbf{Clarity of part boundaries} – Are the decomposed part regions cleanly separated with well-defined borders?
    \item \textbf{Consistency across views} – Do the decomposed parts remain stable and coherent when viewed from different angles?
    \item \textbf{Suitability for rigging/animation} – Are the decomposed parts appropriate for assigning joints and performing realistic articulated motion?
\end{itemize}

Each criterion was rated on a scale from 1 (Poor) to 5 (Excellent).

\paragraph{Goal.}
The objective of this study is to assess the effectiveness of part decomposition methods in real-world rigging scenarios, providing insight into their strengths and limitations for downstream animation applications.

\section{Additional Discussion}

\paragraph{Dual-branch architecture design.}
We investigate both single-branch and dual-branch architectures for jointly predicting multi-view RGB sequences and kinematic part sequences. 
In the single-branch variant, the two modalities are concatenated into a shared latent representation and split prior to decoding. 
This configuration exhibits slower convergence and lower performance than the dual-branch counterpart under the same training schedule. 
We attribute this to the fundamentally different nature of the tasks: RGB synthesis focuses on high-frequency appearance modeling, whereas kinematic part segmentation emphasizes structural reasoning and temporal-spatial consistency. 
A single-branch network that forces both tasks to share all intermediate features is prone to cross-task interference, particularly degrading consistency in multi-frame outputs.

Our BiDiFuse dual-branch UNet addresses this issue by maintaining task-specific feature streams while enabling bidirectional cross-modal feature exchange. 
This architecture preserves modality-specific learning, reduces interference, and improves overall performance. 
We include a detailed discussion of this design choice and relate it to recent work on unified dense prediction in video diffusion models~\cite{yang2025unified}.

\paragraph{Avoiding category-specific pose or rigging priors.}
While 2D/3D pose and rigging priors encode rich kinematic information, SP4D is deliberately designed without reliance on human- or animal-specific templates to ensure category-agnostic applicability. 
Template-based approaches often generalize poorly to objects with unconventional topology, such as loose clothing, handheld tools (e.g., shields, skis), or non-biological categories (e.g., crabs, robotic arms, mechanical assemblies). 
In our experiments, these priors frequently failed to produce meaningful part structures for such diverse object types.

By contrast, SP4D learns kinematic decomposition directly from 2D multi-view supervision, without assuming a fixed skeleton topology, enabling robust generalization to both natural and synthetic domains. 
Nevertheless, integrating lightweight 2D/3D structural priors into diffusion-based generation remains an interesting direction for future research.

\paragraph{Relation to optimization-based methods.}
We also relate SP4D to prior part-aware rendering approaches~\cite{zhangs3o, zhanglearning, yang2022banmo, noguchi2022watch, yang2021viser}. 
These methods are typically optimization-based pipelines for per-instance 3D reconstruction from multi-view videos, often requiring ground-truth camera poses, complete multi-view coverage of the same object, and extensive per-instance optimization (e.g., 48+ GPU hours on an A100). 
Their kinematic reasoning is constrained by the motion observed in the input video; for example, if a limb remains static throughout, the model may fail to segment it. 
Such methods are not designed for category-level generalization and cannot perform feedforward inference from monocular inputs such as a single image or single-view video.

In contrast, SP4D is a feedforward, category-agnostic generative model capable of producing consistent RGB renderings and kinematic part decompositions within seconds, given only a single image or video. 
While both SP4D and part-aware rendering approaches can output kinematic segmentations, they address fundamentally different problem settings and exhibit markedly different capabilities.

\paragraph{Limitations and future work.}
Our method inherits the camera parameterization design from SV4D~2.0~\cite{yao2025sv4d}, which models only azimuth and elevation with a simple lens model. 
This limits our ability to handle videos with strong perspective distortion or complex camera trajectories. 
Moreover, SP4D is primarily trained under the assumption that each scene contains a single object. 
In scenarios where multiple objects appear simultaneously, the model may struggle to handle all of them at once. 
Extending SP4D to support full 6-DoF camera motion and multi-object scenarios remains a promising direction for future research. 
Additional failure cases are provided as supplementary videos to illustrate these challenges.

\textbf{}


\end{document}